\documentclass{article}
\usepackage{arxiv}
\usepackage[utf8]{inputenc} 
\usepackage[T1]{fontenc}    
\usepackage{hyperref}       
\usepackage{url}            
\usepackage{booktabs}       
\usepackage{amsfonts}       
\usepackage{nicefrac}       
\usepackage{microtype}      
\usepackage{graphicx}
\usepackage{natbib}
\usepackage{doi}
\usepackage{comment}
\usepackage{amsmath,amssymb} 
\usepackage{color}
\usepackage{wrapfig}

\begin{document}
\title{The Familiarity Hypothesis: Explaining the Behavior of Deep Open Set Methods} 

\author{Thomas G. Dietterich \textmd{and} Alexander Guyer\\
	Collaborative Robotics and Intelligent Systems (CoRIS) Institute\\
	Oregon State University\\
	Corvallis, OR 97331 USA\\
	\texttt{tgd@cs.orst.edu} \\
}
\renewcommand{\shorttitle}{The Familiarity Hypothesis}
\maketitle

\begin{abstract}
In many object recognition applications, the set of possible categories is an open set, and the deployed recognition system will encounter novel objects belonging to categories unseen during training. Detecting such ``novel category'' objects is usually formulated as an anomaly detection problem. Anomaly detection algorithms for feature-vector data identify anomalies as outliers, but outlier detection has not worked well in deep learning.  Instead, methods based on the computed logits of visual object classifiers give state-of-the-art performance. This paper proposes the \textit{Familiarity Hypothesis} that these methods succeed because they are detecting the \textit{absence} of familiar learned features rather than the \textit{presence} of novelty. This distinction is important, because familiarity-based detection will fail in many situations where novelty is present. For example when an image contains both a novel object and a familiar one, the familiarity score will be high, so the novel object will not be noticed. The paper reviews evidence from the literature and presents additional evidence from our own experiments that provide strong support for this hypothesis. The paper concludes with a discussion of whether familiarity-based detection is an inevitable consequence of representation learning.
\end{abstract}


\section{Introduction}
The open set problem arises in any object recognition system that operates in an open world where novel object classes can arise \citep{Bendale2015}. For example, classifiers for identifying species cannot be trained on all possible species, because new species are discovered every year. Classifiers for recognizing vehicles on roadways encounter novel kinds of vehicles (e.g., hover boards, electric unicycles, roller skis), because new vehicles are invented and marketed frequently. To operate in open worlds, computer vision systems need to master two functions: (a) detecting when an object belongs to a new category and (b) learning to recognize that new category. This paper focuses on the first problem: novel category detection.

\begin{wrapfigure}{R}{2.2in}
    \centering
    \includegraphics[width=2in]{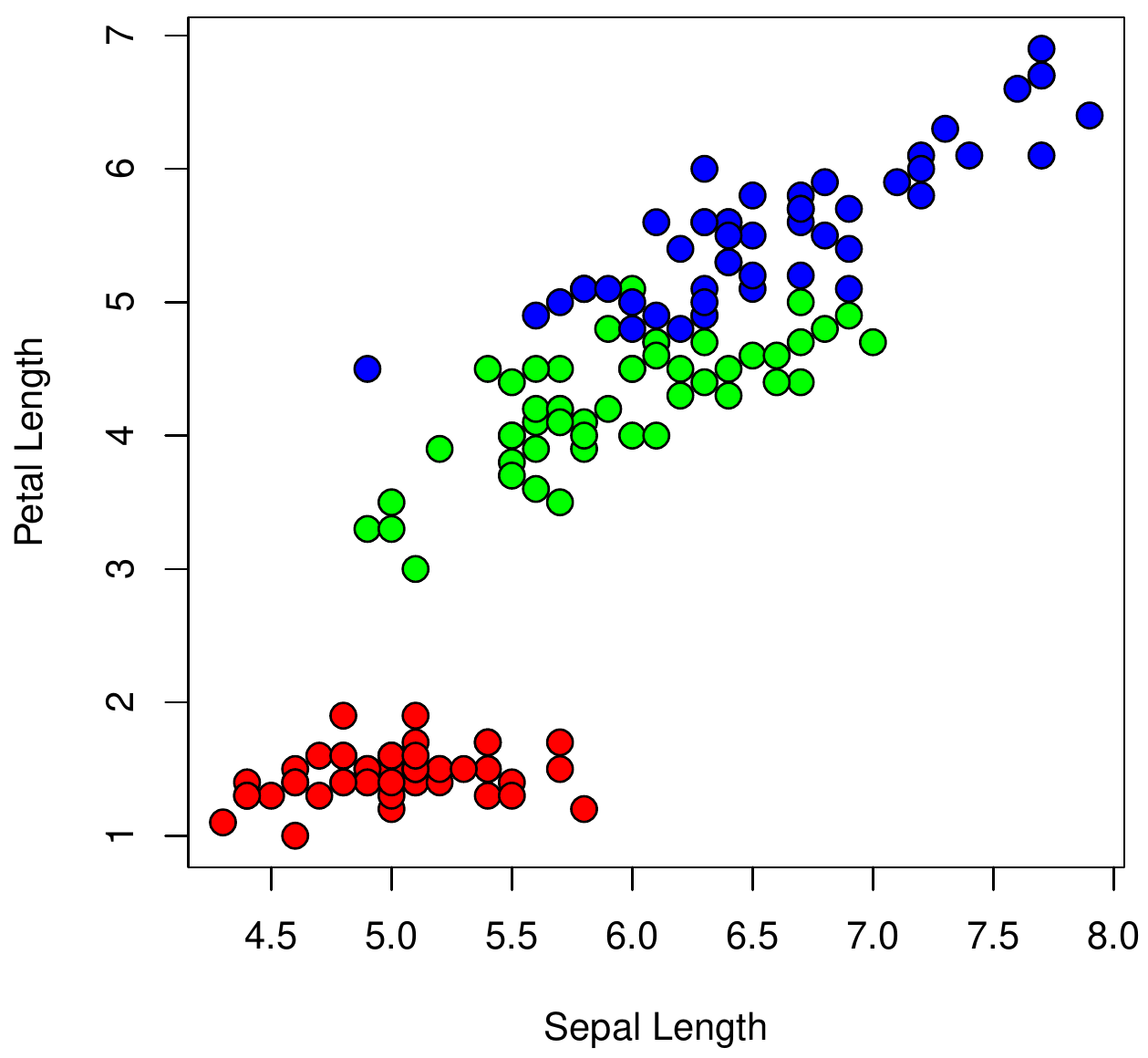}
    \caption{Novel classes are outliers in featurized data. Fisher's Iris data: setosa (red), versicolor (green), virginica (blue).}
    \label{fig:iris}
\end{wrapfigure}
In machine learning applications to feature-vector data, novel category detection has been approached as a problem of outlier detection. The assumption is that a novel query $x_q$ will be an outlier with respect to the training data. For example, Figure~\ref{fig:iris} plots Fisher's famous iris data according to the two features \textit{sepal length} and \textit{petal }length. Suppose that the training categories were \textit{versicolor} and \textit{virginica}. After deployment, flowers belonging to the novel category \textit{setosa} would be obvious outliers, because they have much shorter petals. Similarly, if \textit{versicolor} or \textit{virginica} had been the novel categories, the combination of \textit{petal }length and \textit{sepal} length would make it easy to detect a majority of such flowers as outliers. Outlier detection for feature-vector data is a mature field with many effective and scalable algorithms \citep{Chandola2009,Markou2000,Markou2003,Pimentel2014,Hodge2004,Ruff2021}. These are based primarily on distances \citep{Byers1998,Eskin2002,Liu2012}, density estimates \citep{Kim2012,Pevny2015}, and density quantiles \citep{Scholkopf2001,Tax2004}. 

Many attempts have been made to apply similar ideas in deep learning \citep{Bendale2016,Ruff2018}. Two of the most successful methods are the Mahalanobis method \citep{Lee2018} and OpenHybrid \citep{Zhang2020}. Let $z=E(x)$ denote the vector of activations in the penultimate layer of a deep backbone network $E$ applied to an input image $x$. The Mahalanobis method first trains the backbone using a standard softmax classifier over the known classes. Then it fits a multivariate Gaussian to the $z$ values for each class under the assumption that the covariance matrix $\mathbf{\Sigma}$ of these activations is the same across all classes. Given a query image $x_q$, it computes the squared Mahalanobis distance $(z_q - \hat{\mathbf{\mu}}_k)^\top\mathbf{\Sigma}^{-1}(z_q - \hat{\mathbf{\mu}}_k)$ from $z_q = E(x_q)$ to the estimated mean $\hat{\mathbf{\mu}}_k$ for each class $k\in \{1,\ldots,K\}$. The distance to the nearest class mean provides the anomaly score, with larger scores indicating more extreme anomalies. 

The OpenHybrid method attaches two heads to $E$. One head is a standard softmax classifier, and the other is a flow-based density estimator $P(z)$. The negative log density, $-\log P(z_q)$ provides the anomaly score. These heads are trained concurrently to minimize a combination of the cross-entropy loss for the classifier and the negative log likelihood of $z$ for the density estimator.

A recent paper by \cite{Vaze2021} concluded that while these methods work reasonably well, their performance can be matched by a well-trained classifier. Specifically, let $\ell_k = w_k^\top z_q$ be the logit score of image $x_q$ on class $k$. Vaze et al.~found that the maximum of these logit scores is a good anomaly score (with smaller scores indicating more extreme anomalies).  We will call this the \textit{Standard Model}, because it is closely related to the baseline developed by \cite{Hendrycks2017} that computes the maximum of the softmax scores (obtained by passing the logits through the softmax transformation). The good performance of the standard model matches our own experience and stimulated us to formulate the familiarity hypothesis.

\begin{quote}
    \textbf{The Familiarity Hypothesis (FH)}: \textit{The standard model succeeds primarily by detecting the \emph{absence} of familiar features in an image rather than by detecting the \emph{presence} of novel features in the image.}
\end{quote}

The FH starts from the observation that deep networks learn a set of feature detectors whose activations $z$ are computed by the encoder $z=E(x)$. Each feature detector $j$ computes a potentially complex function of basic patterns present or absent in the input image. We will be specifically interested in \textit{positive presence features} that return a positive activation when some pattern is present in the image and zero when it is absent. Assume for the moment that all of the learned features are positive presence features. (We will consider more complex cases below.) Given the vector $z$ of activations, the logit weight vector $w_k$ then computes a weighted sum $\sum_j w_{k,j}z_j$. Let's focus on the positive weights. A novel object $x_q$ will have fewer features relevant to \textit{any} of the known classes, so it will tend to activate fewer positive-weighted features, and this will decrease the logit score. Hence, the max logit score will provide a good open category detection score.

If the Familiarity Hypothesis is correct, then there are many cases where the standard model will make mistakes. If an object from a known class is occluded in a way that hides familiar features, the max logit score will decrease, and a false novelty will be detected. Conversely, if an image contains two objects, one familiar and one novel, the max logit score will remain high because all of the familiar features will be detected on the familiar object. The novel object will not be noticed at all. Familiarity-based novelty detection will also be more vulnerable to adversarial attacks. An attack that reduces the maximum logit score could cause false novelty detections, and an attack that introduces familiar features into the background of an image containing a novel object could cause missed novelty detections. At a fundamental level, familiarity-based novelty detection is not measuring the true cause of novelty but rather a side effect---the reduction of familiarity.

The goal of this paper is to explore the Familiarity Hypothesis. We assemble evidence bearing on the FH from a combination of published work and our own experiments. Previous work has shown that the activation vectors $z$ have smaller norm and tend to be clustered around the ``center'' of the feature space. We show that these smaller norms are associated with fewer highly-activated features. Then, we present experiments on PASCAL segmentation data showing that the decreased activation of object-sensitive features accounts for most of the decrease in logit scores. In addition to this direct evidence, we discuss how the FH explains published results on the performance of other anomaly scores including the max softmax probability, Mahalanobis distance, generalized ODIN, and adversarial reciprocal points methods. 

To further test the FH, we formulate three predictions and evaluate them experimentally. The results provide additional evidence in favor of the hypothesis. 

Finally, we ask whether there is any way to develop deep anomaly detection methods that do not rely on detecting the absence of familiarity and suggest some promising research directions.

\section{Evidence for the Familiarity Hypothesis}

\begin{figure}
    \centering
    \includegraphics[width=3in]{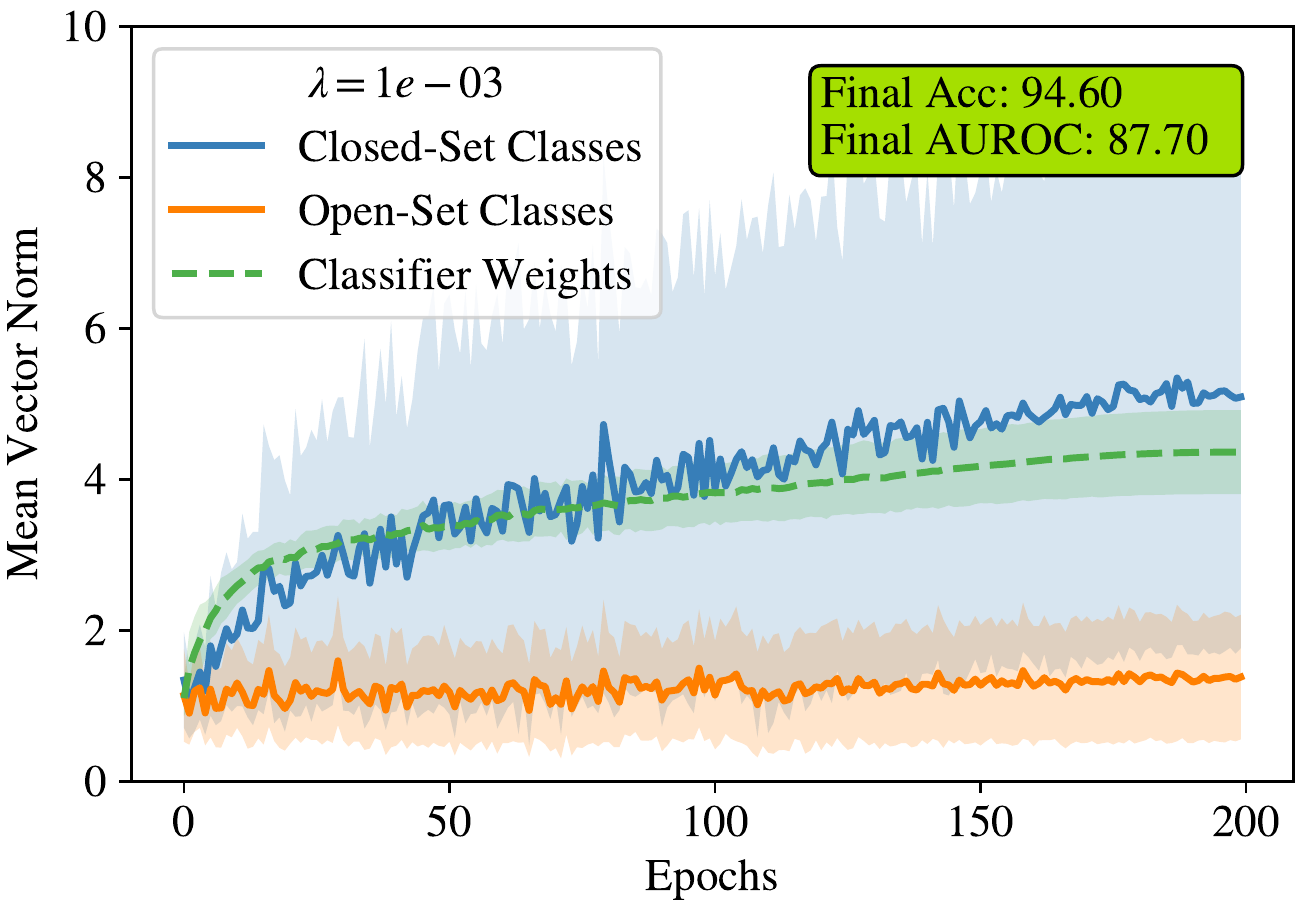}
    \includegraphics[width=3in]{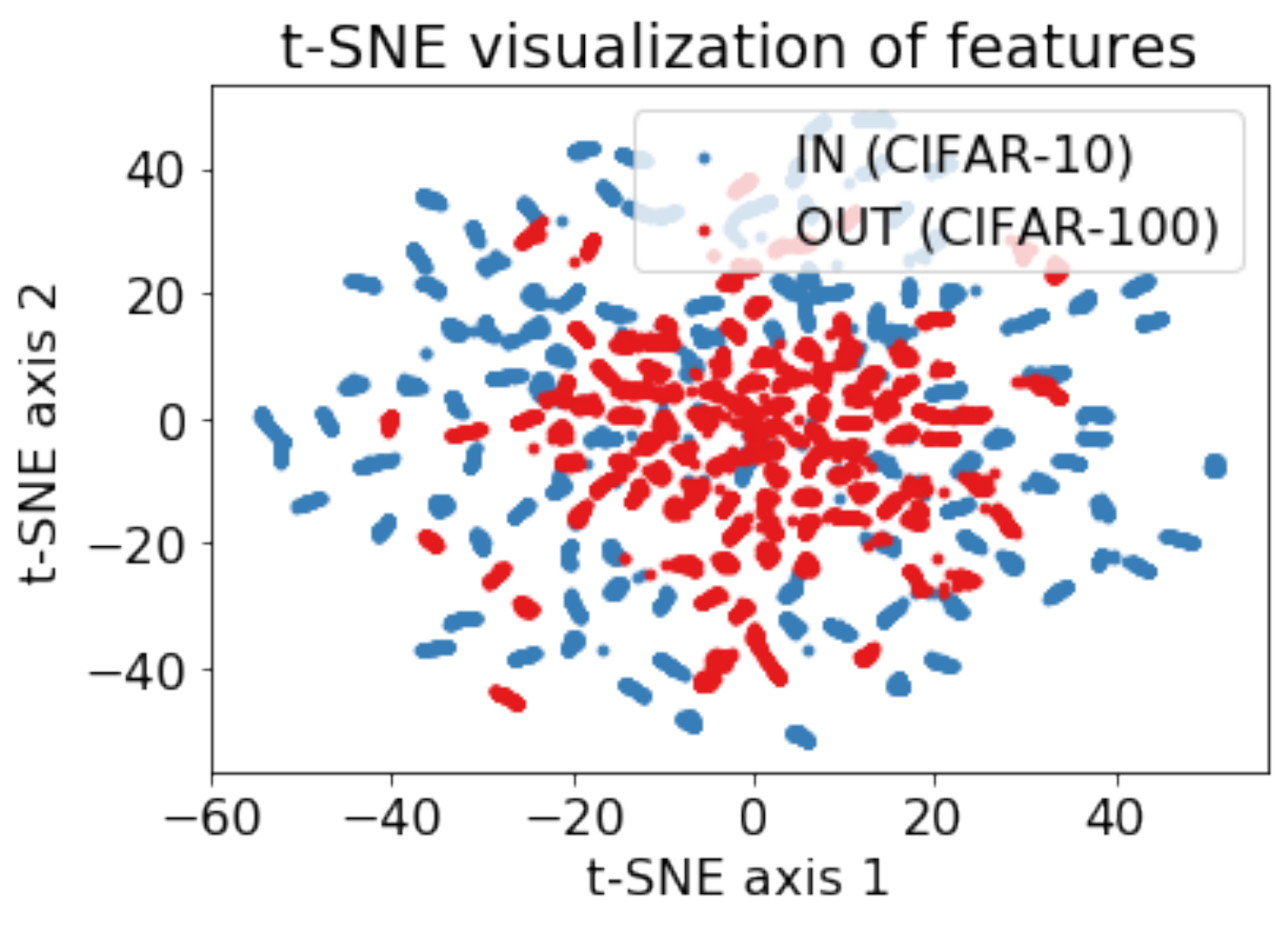}\\
    \includegraphics[width=2.5in]{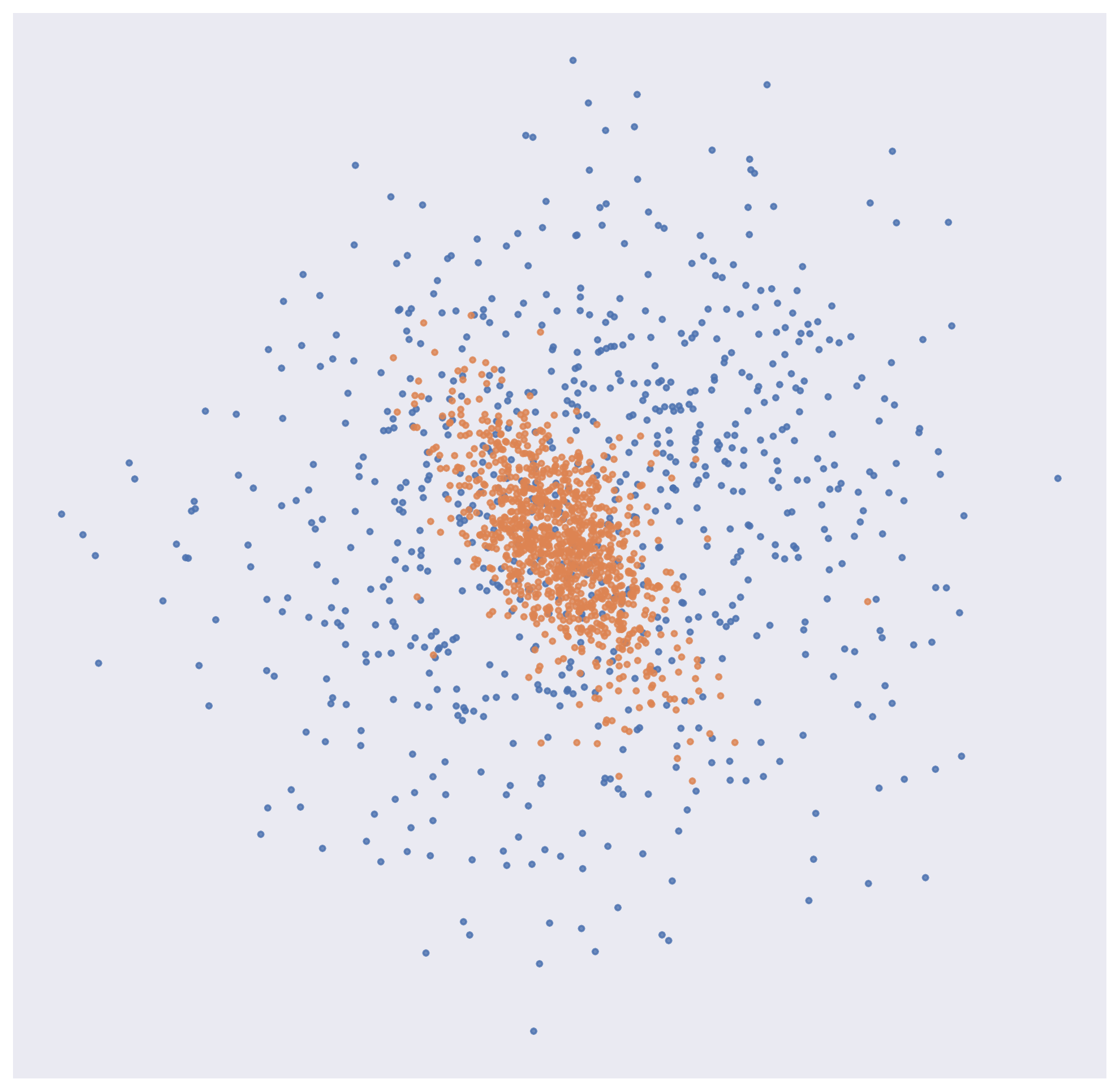}\hspace{0.25in}
    \includegraphics[width=3.5in]{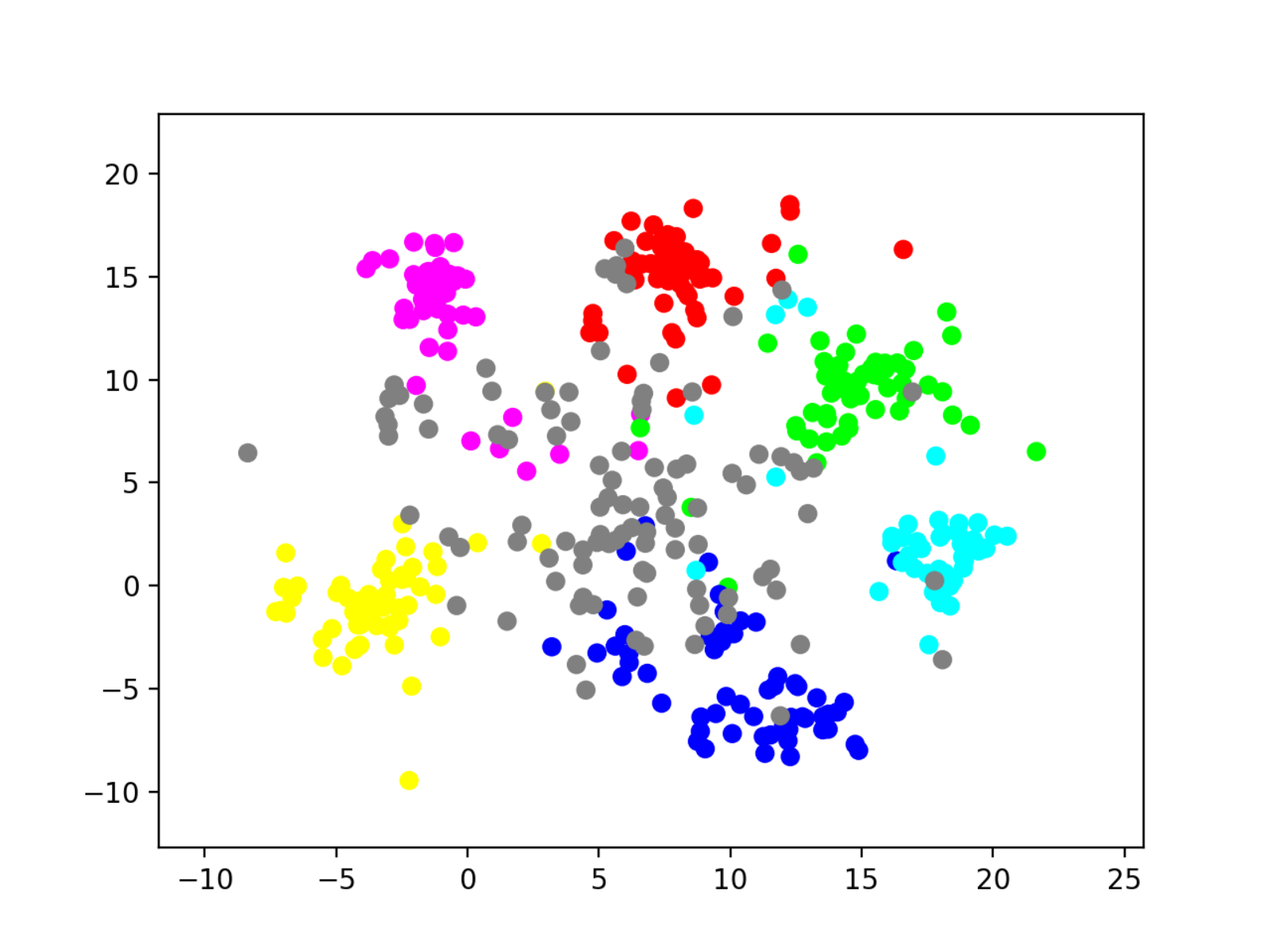}
    \caption{Top left: Norms of $z$ for known and novel classes. Top right: t-SNE visualization of the 10 known classes (blue) and 100 novel classes (red). Bottom left: PCA visualization of MS-1M (known, blue) and ImageNet-1K (novel, orange). Bottom right: UMAP visualization of six known classes (colors) and novel classes (grey).}
    \label{fig:vaze-norm}
\end{figure}
\begin{figure}
    \centering
    \includegraphics[width=3in]{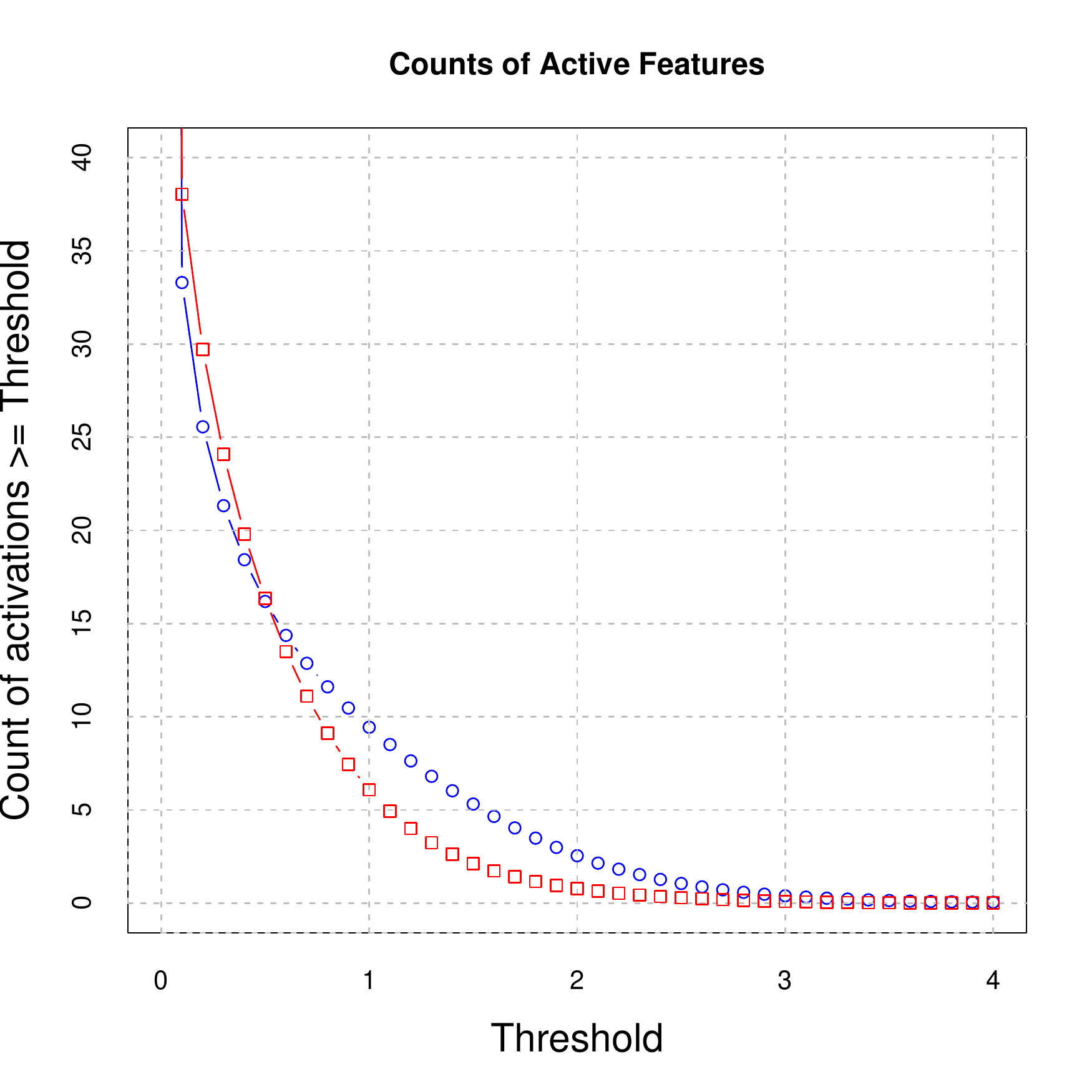}
    \caption{Average number of activations exceeding a threshold as we vary the threshold. Known classes: blue; Novel classes: red.}
    \label{fig:activation-histogram}
\end{figure}
Several authors have discovered that $\|z\|$ for novel class objects is substantially smaller than for objects from known classes. For example, \cite{Vaze2021} trained a VGG32 network on 6 known classes from CIFAR-10 and evaluated on all 10 classes. Figure~\ref{fig:vaze-norm} (top left) plots the mean and standard deviation of $\|z\|$ and the mean classifier weights $\frac{1}{K}\sum_{k=1}^K\|w_k\|$ as a function of the number of epochs. Observe that the $\|z\|$ values for the novel class objects (``open set classes'') are much smaller than for the known class objects (``closed set classes''). This same phenomenon was observed by \cite{Tack2020} training a ResNet18 using instance-contrastive self-supervised learning. What causes this?

Figure~\ref{fig:vaze-norm} (top right), from \cite{Tack2020}, plots a t-SNE \citep{Maaten2008} visualization of the activations of the known class objects (from CIFAR-10) and the novel class objects (from CIFAR-100). The novel class objects cluster closer to the center of the space, so this suggests that the smaller $\|z\|$ results from smaller values along multiple dimensions. Figure~\ref{fig:vaze-norm} (bottom left), from \cite{Huang2021}, plots a PCA visualization of the $z$ values for a network trained on Microsoft's 1M celebrities with novel class objects from ImageNet-1K. Finally, Figure~\ref{fig:vaze-norm} (bottom right), from our own work, plots a UMAP \citep{McInnes2018} visualization of the $z$ values for a DenseNet trained on 6 known classes from CIFAR-10 and tested on all 10 CIFAR-10 classes. Images from known classes are plotted in colors, and images from the four novel classes are plotted in grey. We observe that training has ``pulled'' the known class points away from the center of the space into six nice clusters, whereas the novel class points primarily occupy the center of the space. 

We can obtain a clearer understanding of why the activation vector $z$ has a smaller norm by comparing the activation values against a threshold $\theta$ and counting the number of activations greater than $\theta$, as shown in Figure~\ref{fig:activation-histogram} for our CIFAR-10 6/4 split. For example, consider $\theta=1$. Images of known classes average approximately 10 features with activations above this threshold, whereas novel-class objects average only 6. Except for the smallest activations, the novel class objects always activate fewer above-threshold features than the known classes. 

According to the FH, the features that are activated in images of known objects, and that are not activated in images of novel objects, should be features that describe the object itself rather than the background. To verify this claim, we performed an experiment to measure which features are ``on-object'' versus ``off-object'' (i.e., activated only by the background). We trained a classifier on a subset of the classes in the PASCAL VOC Segmentation Dataset \citep{pascal-voc-2012} and then evaluated this classifier on a mix of known and novel PASCAL classes. To measure which features were activated by the object, we applied a Gaussian blur operator to a copy of the image and then replaced the segmented object in the original image by its blurred version. Features whose activation changes between the unblurred and blurred images are features that must depend (at least partially) on the segmented object region. 

\begin{figure}
    \centering
    \includegraphics[width=\textwidth]{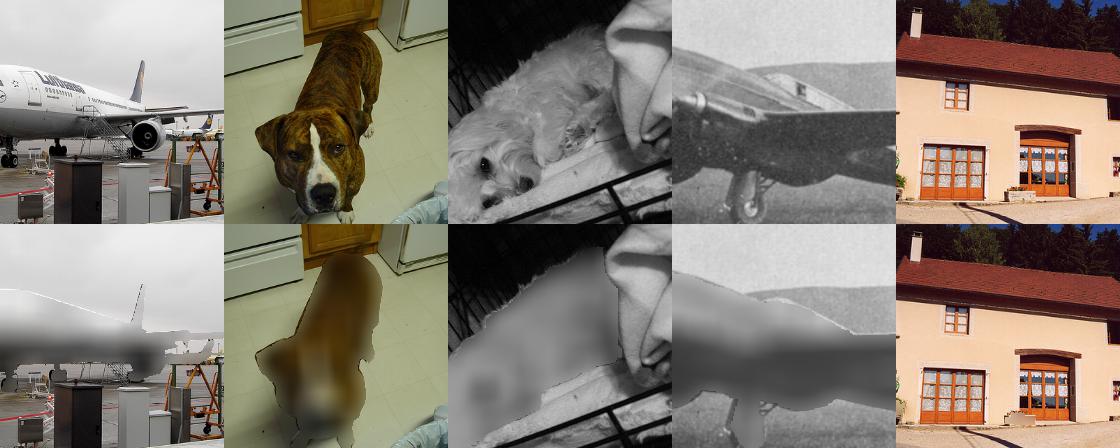}
    \caption{Examples of unblurred images and their blurred counterpairs.}
    \label{fig:blurred-images}
\end{figure}

Images (and segmentation masks) were first resized so that their shortest dimension is 256 pixels. Masks were resized via nearest-neighbor interpolation so that they remained integers. Images were then cropped to 224x224 pixels and a 31x31 Gaussian blurring kernel with standard deviation of 31 was computed over the entire image. Finally, the segmentation mask was applied to replace the original segmented object pixels with the pixels from the blurred image. Note that while this strongly blurs the object, the object boundary remains intact. Hence, learned features that only detect the presence or shape of the boundary may not be affected by the blurring operation. This means that this technique provides sufficient but not necessary conditions for a feature to be ``on-object''. Figure~\ref{fig:blurred-images} shows five examples of unblurred and blurred image pairs. While the blurring does not completely eliminate all of the content and structure of the object, it removes most of it. Even a modest amount of blur changes the activations of features enough to reveal that they are ``on-object'', and our blurring is quite strong.

In order to construct an image-level classification data set where each image has only one class label, we eliminated from consideration all PASCAL images that contained multiple objects. We also ignored objects belonging to the ``person" and ``chair" classes, as they commonly co-occurred with other objects. The remaining dataset contained only 3,794 images, so we increased the size of the training set by selecting a total of 295,434 images from 233 corresponding fine-grained classes in ImageNet-1K. For example, images of dogs of various breeds from ImageNet-1K were merged with the dog-class images in PASCAL VOC. We removed three more PASCAL classes (cows, horses, and sheep) because there were no ImageNet-1K examples available for those classes. Of the remaining 15 classes, we randomly selected eight (``aeroplane'', ``bicycle'', ``bird'', ``boat'', ``diningtable'', ``dog'', ``pottedplant'', and ``train'') to be considered known and held out the remaining seven (``bottle'', ``bus'', ``car'', ``cat'', ``motorbike'', ``sofa'', and ``tvmonitor'') as novel. To alleviate issues with class imbalance during training, for each minibatch, we sampled images randomly using sample weights inversely proportional to the frequency of the image class. Let $N_{id}$ and $N_{ood}$ be the number of known (``in distribution'') and novel (``out of distribution'') images in the test data. 

We processed the activation vectors of the unblurred and blurred in-distribution images to develop a score $OO(j,k)$ that characterizes whether a feature $j$ is ``on-object'' in the images that are predicted to belong to class $k$. In our notation, a tilde indicates a quantity computed on the blurred images. Hence, $\tilde{z}_i$ is the activation vector $E(\tilde{x}_i)$ for the blurred version of image $i$, whereas $z_i = E(x_i)$ is the activation vector of the unblurred image.

Let 
\[BE(i,j) = z_{ij} - \tilde{z}_{ij}
\]
quantify the ``blurring effect'' on feature $j$ in image $i$. If $BE(i,j)$ is positive, this means that blurring reduced the activity of feature $j$ on image $i$. We will say that feature $j$ is acting as a ``presence feature'', because its activity is higher when the feature is present. Conversely, if $BE(i,j)$ is negative, blurring increased the activity of the feature. We will say $j$ is acting as an ``absence feature'' in image $i$, because removing the feature by blurring caused activity to increase. One way for a ReLU network to implement an absence feature is to subtract a positive activation from a large positive bias in one of the intermediate layers. If $BE(i,j) \approx 0$, then blurring had no effect on feature $j$. This might be because feature $j$ is only sensitive to the background or that it is sensitive to the boundary of the object (which blurring does not remove). It could also be that feature $j$ is measuring some combination of presence and absence patterns in the image, and the sum of those measurements is zero on image $i$. Work on explaining deep networks in computer vision has shown that a single feature $j$ can be computing a complex logical combination of several primitive patterns \citep{LiFuxin2021}. The value of $BE(i,j)$ is only measuring the net effect of blurring on the image. 

To measure the overall effect of feature $j$ on images labeled with class $k$, we define an on-object score as the average blurring effect over all $N_k$ images in class $k$:
\[
OO(j,k) = \frac{1}{N_k} \sum_{\{i|y_i = k\}} z_{ijk} - \tilde{z}_{ijk}.
\]
We will say that a feature $j$ is a net presence feature if $OO(j,k)>0.02$, a net absence feature if $OO(j,k)<-0.02$, and a net neutral feature otherwise. The threshold of 0.02 was chosen heuristically to remove features that have little or no impact on the max logit scores. Setting the threshold to 0 has no effect on our conclusions, but the visualizations become messier.

Let us now consider how the activation of a feature contributes to computing the logit of a class. Define $c_{ijk} = w_{jk} z_{ij}$ to be the \textit{contribution} of feature $j$ to the logit for class $k$ on image $i$, and $\tilde{c}_{ijk}$ to be the corresponding value for the blurred image.    
Define the mean contribution of feature $j$ to class $k$ to be $\overline{c}_{jk} = (1/N_{k}) \sum_{\{i|y_i = k\}} c_{ijk}$ and $\overline{\tilde{c}}_{jk}$ be the corresponding blurred quantity. 

\begin{figure}
    \centering
    \includegraphics[width=\textwidth]{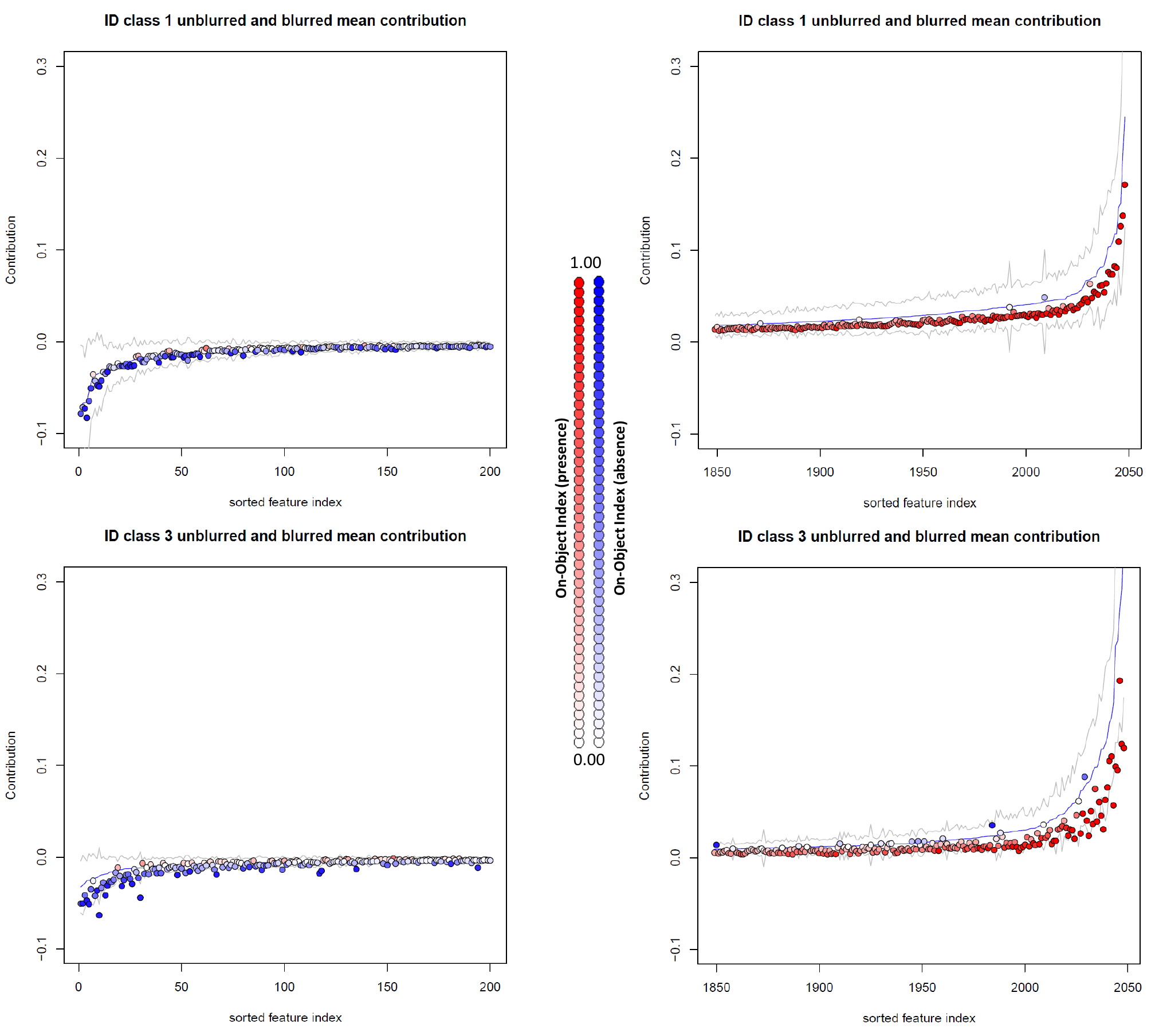}
    \caption{Relevant mean contributions of features unblurred (blue line $\pm$ 1 s.d. in grey) and blurred (points). Point color encodes scaled magnitude of on-object index.}
    \label{fig:mean-sun-li}
\end{figure}

Figure~\ref{fig:mean-sun-li} plots the values of $\overline{c}_{jk}$ (as the blue line) and $\overline{\tilde{c}}_{jk}$ (as the colored points) for PASCAL known classes $k=1$ and $k=3$. The values are sorted in increasing order of $\overline{c}_{jk}$, and we have focused on the 200 smallest and 200 largest values, as the remaining values are all near zero. The grey lines show $\pm$ one standard deviation about $\overline{c}_{jk}$. The color in the colored points encodes $OO(j,k)$ (positive values in shades of red and negative values in shades of blue). Specifically, the color represents a linear interpolation between 0 and the 98th percentile of the observed magnitudes of the positive (and, conversely, negative) values for class $k$. We trim the extreme 2\% of the data points of this distribution to make this visualization robust to outliers. 

\begin{table}
    \centering
    \caption{Taxonomy of four feature types}
    \label{tab:feature-types}
    \begin{tabular}{l|c|c}
                & $w_{jk}>0$        & $w_{jk}<0$        \\\midrule
    $OO(j,k)>0.02$ & positive presence & negative presence \\
    $OO(j,k)<-0.02$ & positive absence  & negative absence 
\end{tabular}
\end{table}

For each class $k$, the features can be partitioned into four types based on the sign of $OO(j,k)$ and the sign of $w_{jk}$, as shown in Table~\ref{tab:feature-types}. Positive presence features decrease in activation when the object is blurred, and they have a positive weight for class $k$. Hence, these features cause the logit to decrease when the image is blurred. Negative absence features increase in activation when the object is blurred, and they have a negative weight. Hence, these also cause the logit to decrease when the object is blurred. Conversely, negative presence features and positive absence features cause the logit score to increase when the object is blurred. A more refined statement of the Familiarity Hypothesis is that most of the change in the max logit score will be caused by positive presence and negative absence features, because these are the familiar features that the classifier is expecting to be present and absent. The negative presence and positive absence features should have only a small effect on the max logit score.

Returning to Figure~\ref{fig:mean-sun-li}, note that for both classes 1 and 3, the blue points in the left figure are negative absence features and the red points in the right figure are positive presence features. These show that on average, blurring will decrease the logit score. There are a few blue points in the right-side figures and a few red points in the left-side figures. These are features that, when blurred, will act to increase the logit score.

\begin{figure}
    \centering
    \includegraphics[width=3in]{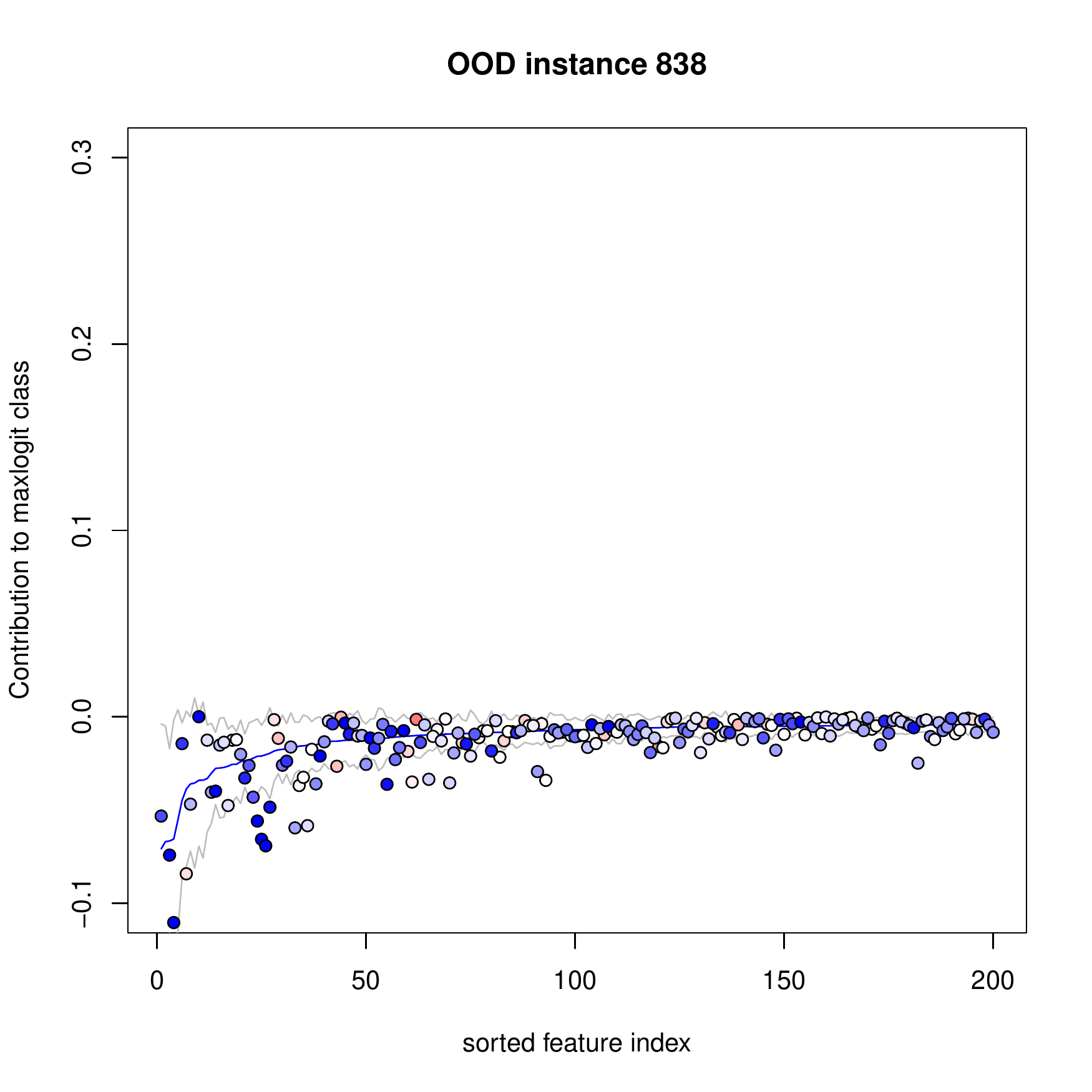}
    \includegraphics[width=3in]{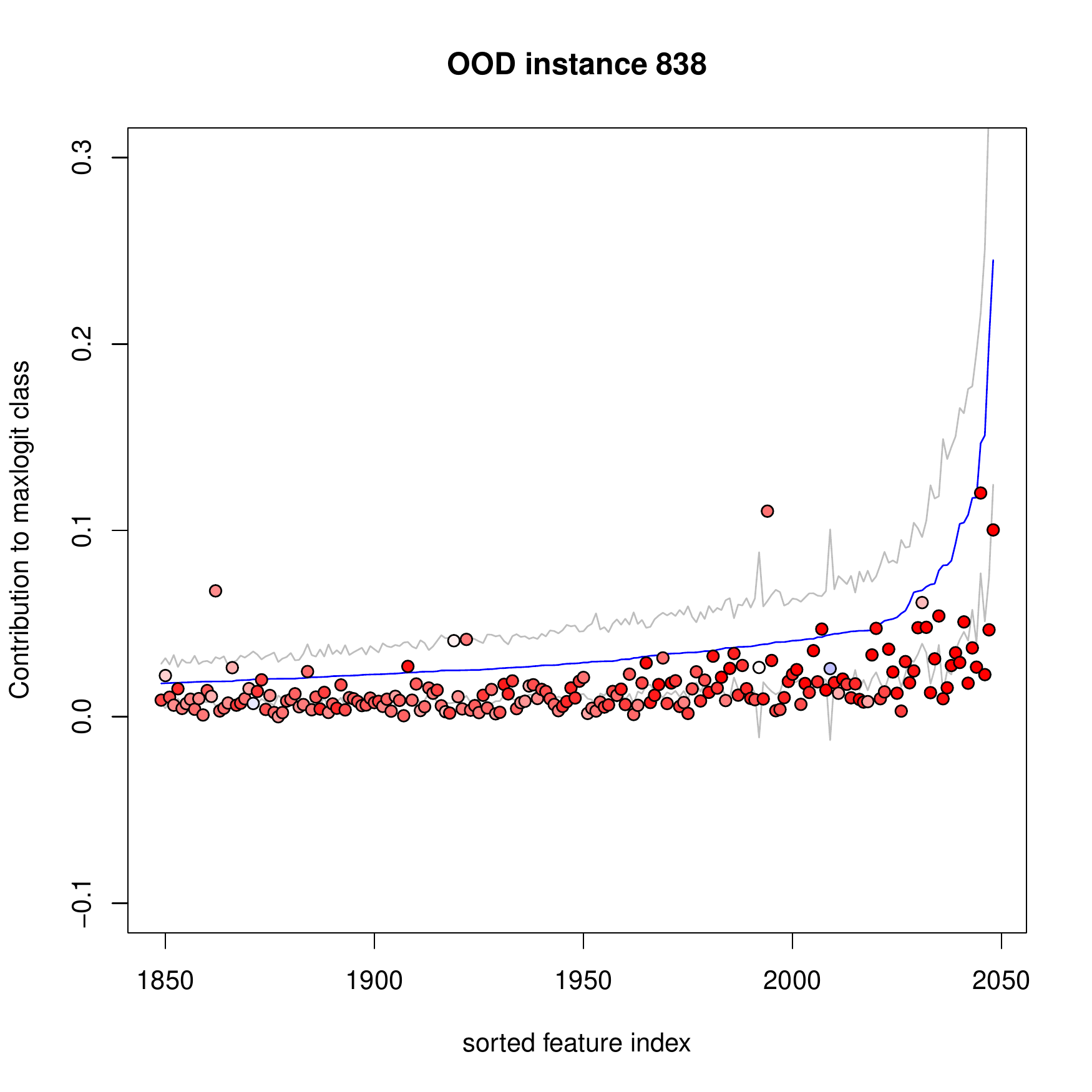}
    \caption{Contribution of relevant features to the max logit score for OOD instance 838. 
    Blue line is mean contribution of ID features to the logit and grey lines show $\pm 1$ s.d. Points are the contributions of instance 838 to the logit. Point color encodes the sign and scaled magnitude of on-object index.}
    \label{fig:ood-instance-838}
\end{figure}

The familiarity hypothesis claims that novel objects will have much the same effect as blurring. We can confirm this by examining the contributions of each feature on individual OOD images. Figure~\ref{fig:ood-instance-838} shows the results for image 838, which is the image with the highest novelty score (smallest max logit). The colored dots plot the actual contributions $c_{ijk}$ of the features for this one image. Observe that most of the dots in the right figure are red and lie far below the mean contribution line. These are positive presence features. The situation in the left figure is more complex. There are many blue dots lying below the mean contribution line, and these are negative absence features that are also acting to reduce the logit score. But there are also many features that, on average, are negative absence features but whose activations decreased in image 838. Hence, they are contributing to increase the logit score. The overall qualitative observation is that most of the departure from the mean contributions is due to positive presence and negative absence features---exactly as predicted by the Familiarity Hypothesis. 

We can summarize the change in the expected logit score in terms of these four feature types. Let $\delta_{ijk} = \overline{c}_{jk} - c_{ijk}$ quantify the decrease in the class $k$ logit score caused by feature $j$ of image $i$. For image 838, the sum of $\delta_{ijk}$ values for positive presence features is 6.232. We will call this the ``positive presence effect.'' The sum of $\delta_{ijk}$ values for negative absence features (the ``negative absence effect'') is 1.114. For the positive absence and negative presence features, the corresponding values are 0.050 and 0.130, which are still positive, but very close to zero. This confirms the Familiarity Hypothesis. 

\begin{figure}
    \centering
    \includegraphics[width=3in]{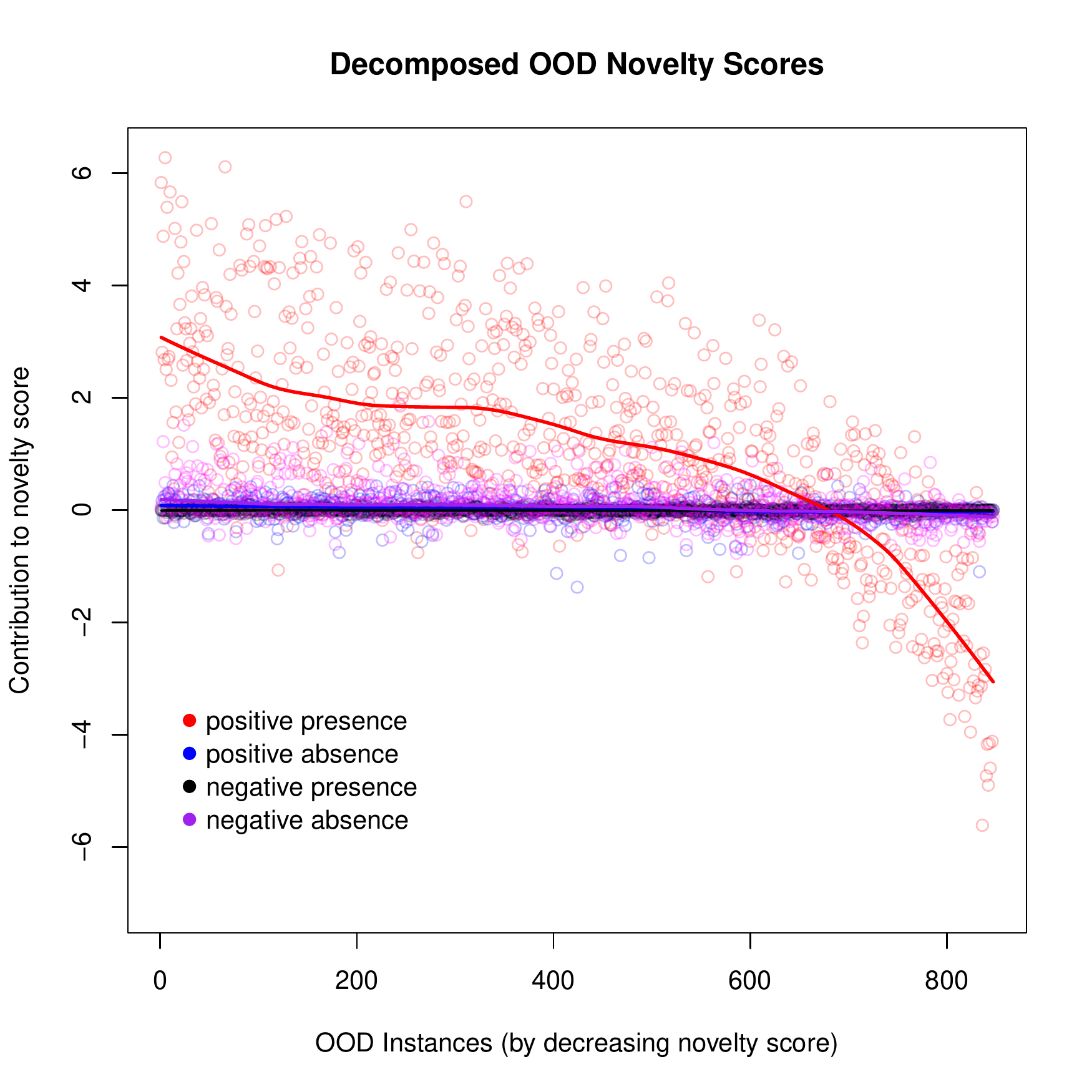}
    \includegraphics[width=3in]{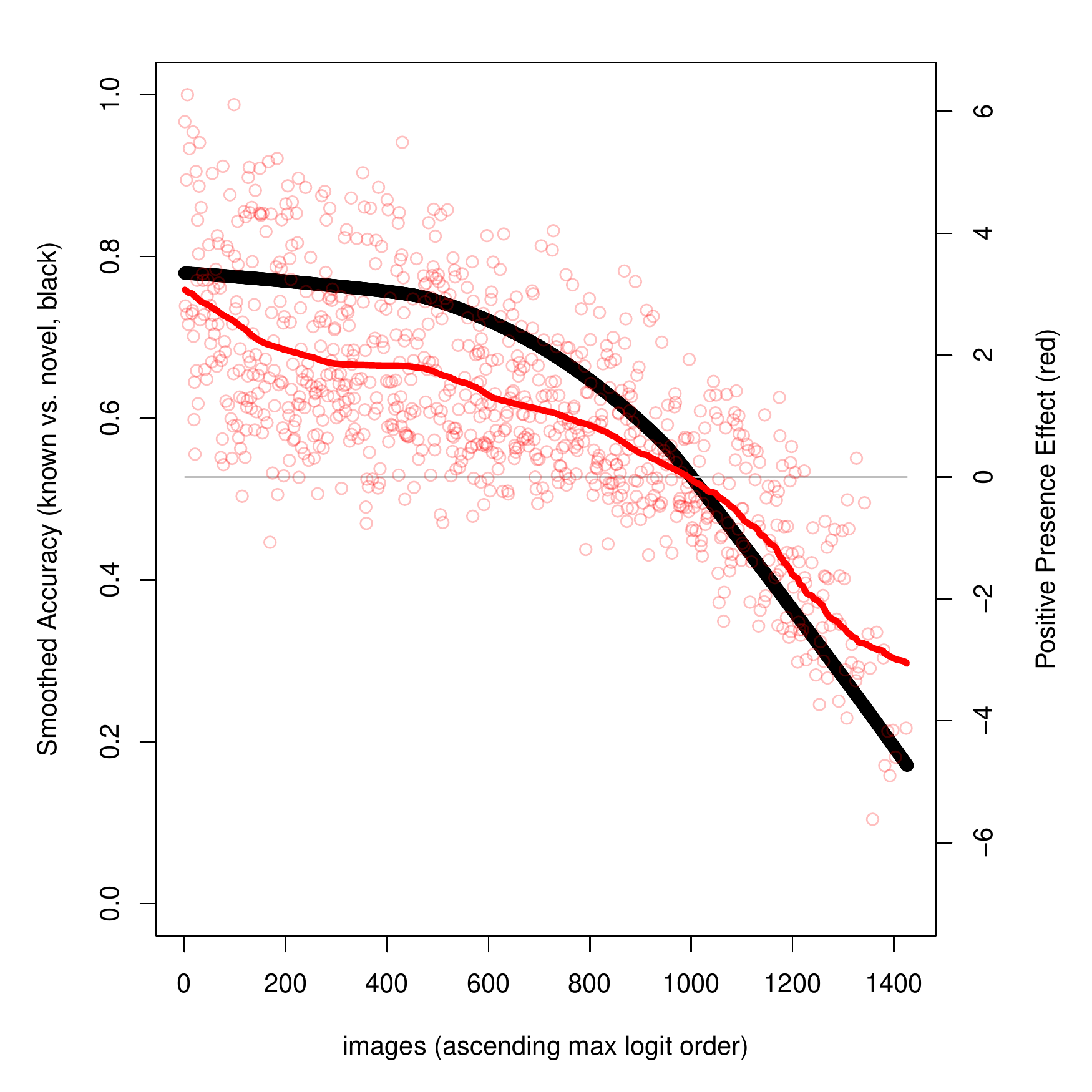}
    \caption{Left: Decomposed novelty scores. Lines are R LOWESS fits to the points with $f=0.25$. Right: Anomaly detection accuracy (black, left axis) and positive presence effect (red, right axis).}
    \label{fig:decomposed-contributions}
\end{figure}

Figure~\ref{fig:decomposed-contributions} (left) plots these four effects for all 857 OOD images (sorted in increasing max logit order). For each image, four dots are plotted, corresponding to the four effects for that image. We observe that the positive presence features are responsible for virtually all of the variation in novelty score. These features make positive contributions for the OOD images with the strongest novelty signal, but the contributions become negative for the OOD images with the smallest novelty signals. Similar, but very weak, trends are observed for the ``negative absence'' and ``positive absence'' features. The ``negative presence" features contribute almost nothing to the score. This confirms the FH that it is the decreased activations of the positive presence features that are having the biggest effect on the max logit scores. In these OOD images, the learned network is not detecting the features that normally make positive contributions to the logit scores.

Figure~\ref{fig:decomposed-contributions} (right panel) replots the positive presence effects for each point. The overlaid black line is a smooth estimate (computed using LOWESS; \cite{Cleveland1981}) of the anomaly detection accuracy. We see that this directly tracks the contributions of the positive presence effects. This suggests that to improve the anomaly detection accuracy, it is important to learn more specific on-object features that are not activated by objects belonging to novel classes. 

The reader should note that because the softmax loss is invariant to additive changes in logit scores ($\ell_k + b$ for all $k$), the definition of positive and negative weights depends on exactly how the neural network is trained. Because we initialized the network weights to small values near zero and because weight decay is applied throughout training, the logit weights are almost perfectly centered around zero. To be precise, $\sum_k w_{jk} \approx 0$ for all $j$. Hence, features where $w_{jk} < 0$ act to reduce the logit value for class $k$ relative to the other classes, and features where $w_{jk} > 0$ increase the relative logit value for class $k$. 

\section{Predictions Based on the Familiarity Hypothesis}
\def\prediction#1{\noindent\textbf{#1}}

To further test the Familiarity Hypothesis, we make the following predictions.

\prediction{Prediction 1: Max Logit Will Give Better Novelty Detection than the Mahalanobis Method.} The Mahalanobis method \citep{Lee2018} considers all activations in $z$ to be potentially relevant for class $k$. Because the Mahalanobis method fits a shared covariance matrix, the effective weight it assigns to a feature depends only on the variability of that feature across the entire training set (and its covariance with other features), so it is learning a single weighted metric for all classes. In contrast, the Max Logit method only considers features that the classifier has determined are relevant on a class-by-class basis. Hence, the FH predicts that Max Logit scores will give better novel detection signals than Mahalanobis.

\prediction{Prediction 2: Object Occlusions will Cause False Positive Novelties.}
If a known object is substantially occluded in an image, this will cause its logit score to shrink, and that could lead to a false positive novelty score. 

\prediction{Prediction 3: Training on Auxiliary Classes will Improve Classification and Novelty Detection Performance.} In most applications of computer vision, there is a set of classes $\mathcal{Y}_{known}$ that constitute the performance task. However, we may also have access to image data labeled with additional classes, which we will refer to as $\mathcal{Y}_{aux}$, for ``auxiliary''. We predict that training on the union $\mathcal{Y}_{known} \cup \mathcal{Y}_{aux}$ will create features that are more informative for the known classes and therefore, according to FH, these will produce more accurate novelty detection signals. 

\section{Experimental Tests of Our Predictions}

We conducted a series of experiments to test these predictions.

\subsection{Comparison of Max Logit with Mahalanobis and other scoring methods}
We performed open set detection experiments to compare four anomaly signals---Max Logit, Mahalanobis, Max Softmax probability, and Directed Sparsification (DICE) \citep{Sun2021}---on five data sets---CIFAR-10 (6/4 split; \cite{Krizhevsky2009}), CIFAR-100 (10/10 split; \cite{Krizhevsky2009}), ImageNette \citep{Howard2020}, and Imagenet-1K with two different sets of 1000 novel classes, as defined by \cite{Vaze2021}. Table~\ref{tab:method-comparison} reports the area under the ROC curve for the binary classification task of detecting image classes as known versus novel. The standard deviations are computed over five replications, each with a separate random split between the known and novel classes (except for Imagenet-1K, where only one split is provided by \cite{Vaze2021}). The results confirm that the Max Logit score gives the best performance, with Max Softmax and DICE being somewhat worse. The very poor performance of the Mahalanobis method confirms Prediction 1.

\begin{table}
    \centering
    \caption{Comparison of anomaly score AUROC for four scoring methods on five data sets}
    \label{tab:method-comparison}
  \begin{tabular}{@{}llllll@{}}\\\toprule
Data set      & CIFAR-10        & CIFAR-100       & ImageNette      & ImageNet-1K & ImageNet-1K \\
              &                 &                 &                 & (Easy)   & (Hard)   \\\midrule 
Anomaly       & Mean AUC        & Mean AUC        & Mean AUC        & AUC         & AUC         \\
Score         & $\pm$ s.d.      & $\pm$ s.d.      & $\pm$ s.d.      &             &             \\\midrule
Max Logit     & 0.883$\pm$0.035 & 0.757$\pm$0.033 & 0.814$\pm$0.029 & 0.8181      & 0.7317      \\
Mahalanobis   & 0.597$\pm$0.042 & 0.540$\pm$0.049 & 0.454$\pm$0.070 & 0.6347      & 0.5150      \\
Max Softmax   & 0.878$\pm$0.026 & 0.726$\pm$0.034 & 0.784$\pm$0.030 & 0.7854      & 0.7183      \\
DICE          & 0.868$\pm$0.041 & 0.738$\pm$0.047 & 0.771$\pm$0.023 & 0.8147      & 0.7327      \\\midrule
Known Classes & 6               & 10              & 6               & 1000        & 1000        \\
Novel Classes & 4               & 10              & 4               & 1000        & 1000        \\
Replications  & 5               & 5               & 5               & 1           & 1           \\
Architecture  & DenseNet 100    & DenseNet 100    & ResNet 34       & ResNet 34   & ResNet 34   \\ \bottomrule
  \end{tabular}
\end{table}

We included the DICE method because \cite{Sun2021} reported substantial increases in AUROC performance on out-of-distribution (cross-data set) experiments. DICE sets 90\% of the logit weights to zero based on the value of the mean contributions. This includes all of the negative weights and many small positive weights. It retains only those weights corresponding to the top 10\% most positive mean contributions $\overline{c}_{jk}$. DICE then employs the denominator of the softmax as the anomaly score. The results here suggest that DICE provides little or no gain in open set performance. Sun \& Li also reported that the Mahalanobis method gives poor out-of-distribution performance.

\subsection{Occlusion}
Blurring the object in an image can be viewed as an extreme form of occlusion. Hence, our blurring manipulation of the PASCAL VOC data provides a simple way to test Prediction 2 regarding occlusion. We compared the AUROC for two settings: (a) unblurred images of known class objects versus unblurred images of novel class objects and (b) blurred images of known class objects versus unblurred images of novel class objects. The corresponding AUROC values are 0.671 (95\% confidence interval of [0.642, 0.700]) and  0.540 (95\% confidence interval of [0.509, 0.571]). Clearly, occluding the principal object leads to a dramatic loss in AUC. Depending on where we set the anomaly detection threshold, this would result in very low true positive rates or very high false positive rates. This supports Prediction 2.
\begin{table}
\centering
\caption{AUROC with Normal and Auxiliary training}
\label{tab:cifar-100}
\begin{tabular}{@{}ll@{}}\\\toprule
Configuration & AUC $\pm$ s.d.    \\ \midrule
Normal        & 0.757 $\pm$0.033 \\
Auxiliary     & 0.886 $\pm$0.036 \\ \bottomrule
\end{tabular}
\end{table}
\begin{figure}
    \centering
    \includegraphics[width=3in]{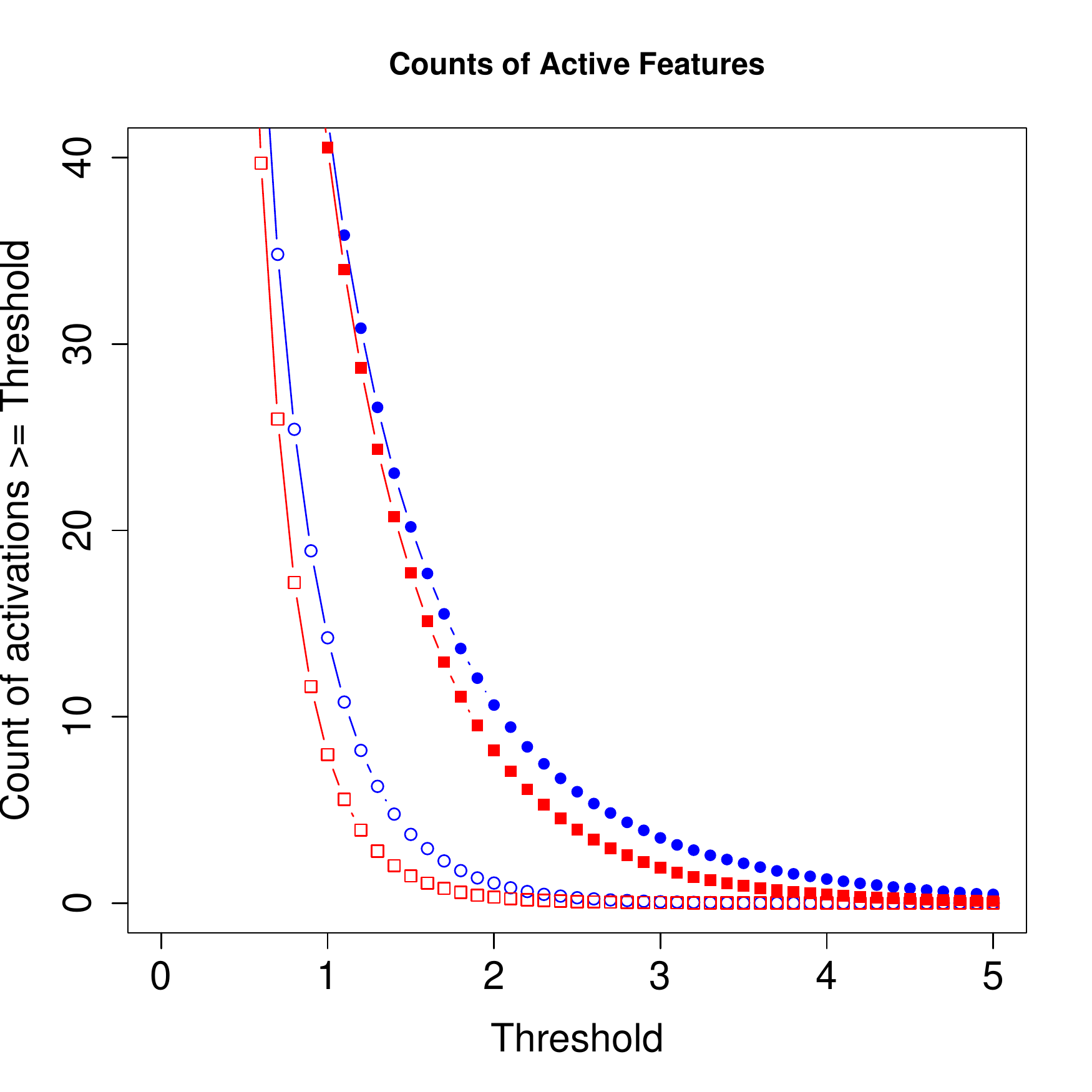}
    \caption{Number of feature activations exceeding a threshold as we vary the threshold. Blue: known-class images; Red: novel-class images; Open shapes: Normal training; Filled shapes: Auxiliary training}
    \label{fig:cifar-100-activations}
\end{figure}

\subsection{Training on Auxiliary Classes}
To study Prediction 3, we performed the following experiments using the CIFAR-100 data set. We chose 10 classes at random to serve as the known classes and a second set of 10 classes to serve as the novel classes. The labeled training data from the remaining 80 classes was then employed as auxiliary training data. We compared two configurations. In the ``Normal'' configuration, we trained a DenseNet100 on only the 10 known classes, whereas in the ``Auxiliary'' configuration, we trained a DenseNet100 on 90 classes (10 known and 80 auxiliary). In both cases, we minimized the softmax cross-entropy loss. We performed five replications of this procedure. The anomaly score is computed as the max logit over the 10 known classes---the 80 auxiliary classes are ignored during anomaly scoring. Table~\ref{tab:cifar-100} shows that auxiliary training greatly improves the anomaly detection performance of the learned network. 

Earlier, we saw in Figure~\ref{fig:activation-histogram} that novel-class images produced smaller activations on CIFAR-10. Figure~\ref{fig:cifar-100-activations} replicates this analysis on both the Normal and Auxiliary configurations. The plot reports the mean number of features whose activation exceeds a threshold $\theta$ as the threshold is varied from 0 to 5. Open shapes show values for Normal training, whereas filled shapes show values for Auxiliary training. Blue circles indicate known-class values, and red squares indicate novel-class values. Both Normal and Auxiliary training exhibit the same phoenomenon: The novel-class images systematically activate fewer features. Note that the magnitude of the activations is substantially larger with Auxiliary training. We believe this is a consequence of the cross-entropy loss, which requires that the maximum logit value for the target class be very high in order to drive the predicted probability for that class close to 1.0 (and, hence, the cross-entropy loss close to zero).

\begin{figure}
    \centering
    \includegraphics[width=3in]{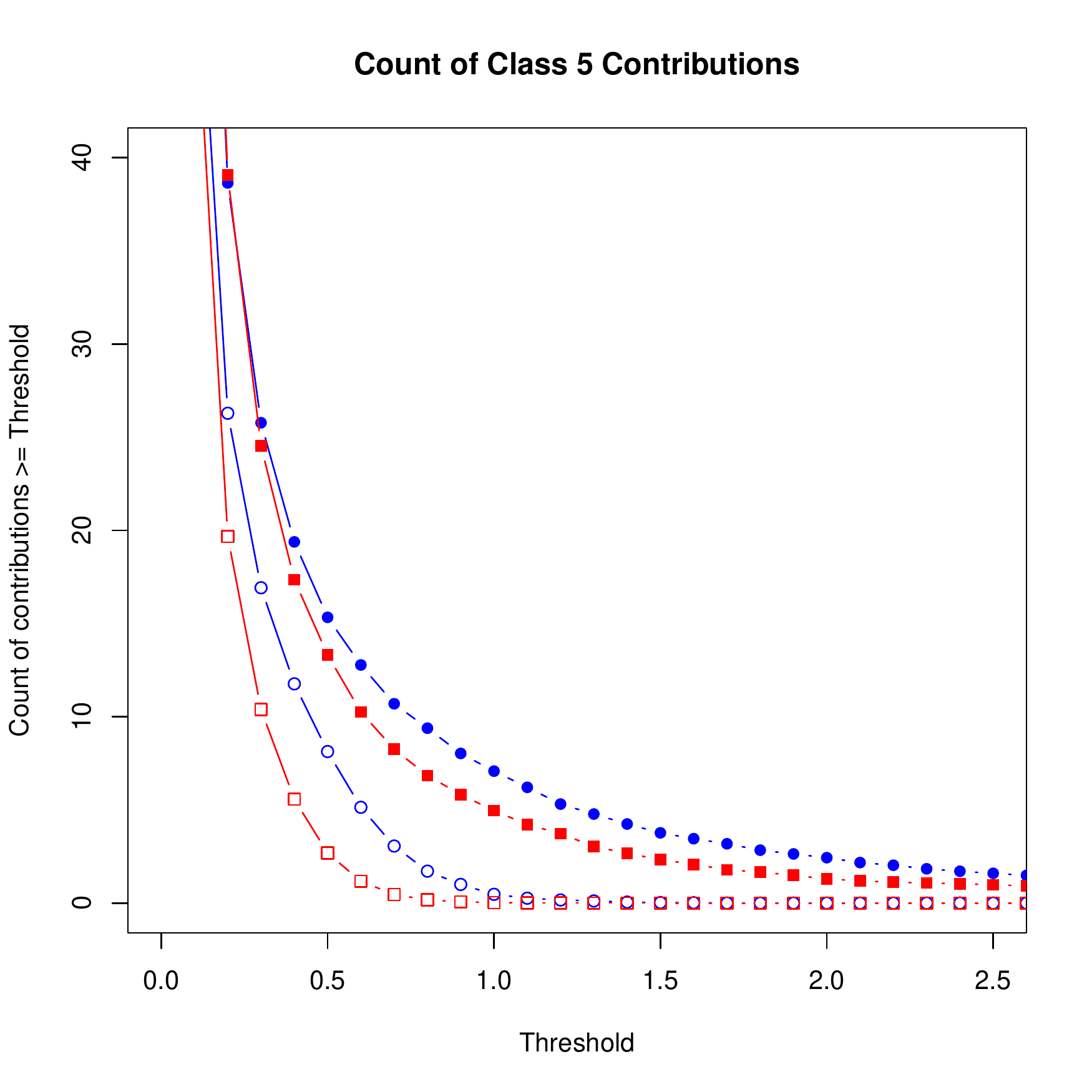}
    \caption{Number of feature contributions to class 5 exceeding a threshold as we vary the threshold. Blue: known-class images; Red: novel-class images; Open shapes: Normal training; Filled shapes: Auxiliary training}
    \label{fig:cifar-100-class-5-contributions}
\end{figure}

To gain further understanding, we applied the same analysis to the absolute values of the contributions $|c_{ijk}|$ for instances $i$ that are predicted by the network to belong to class $k$.  Figure~\ref{fig:cifar-100-class-5-contributions} plots the number of contributions to class 5 whose absolute value exceeds a threshold $\theta$ as the threshold is swept from 0 to 2.6. We observe that in both Normal and Auxiliary training, the contributions produced by novel-class images are smaller than the contributions produced by known-class images. Plots for the other 9 known classes show the same patterns.

\begin{figure}
    \centering
    \includegraphics[width=3in]{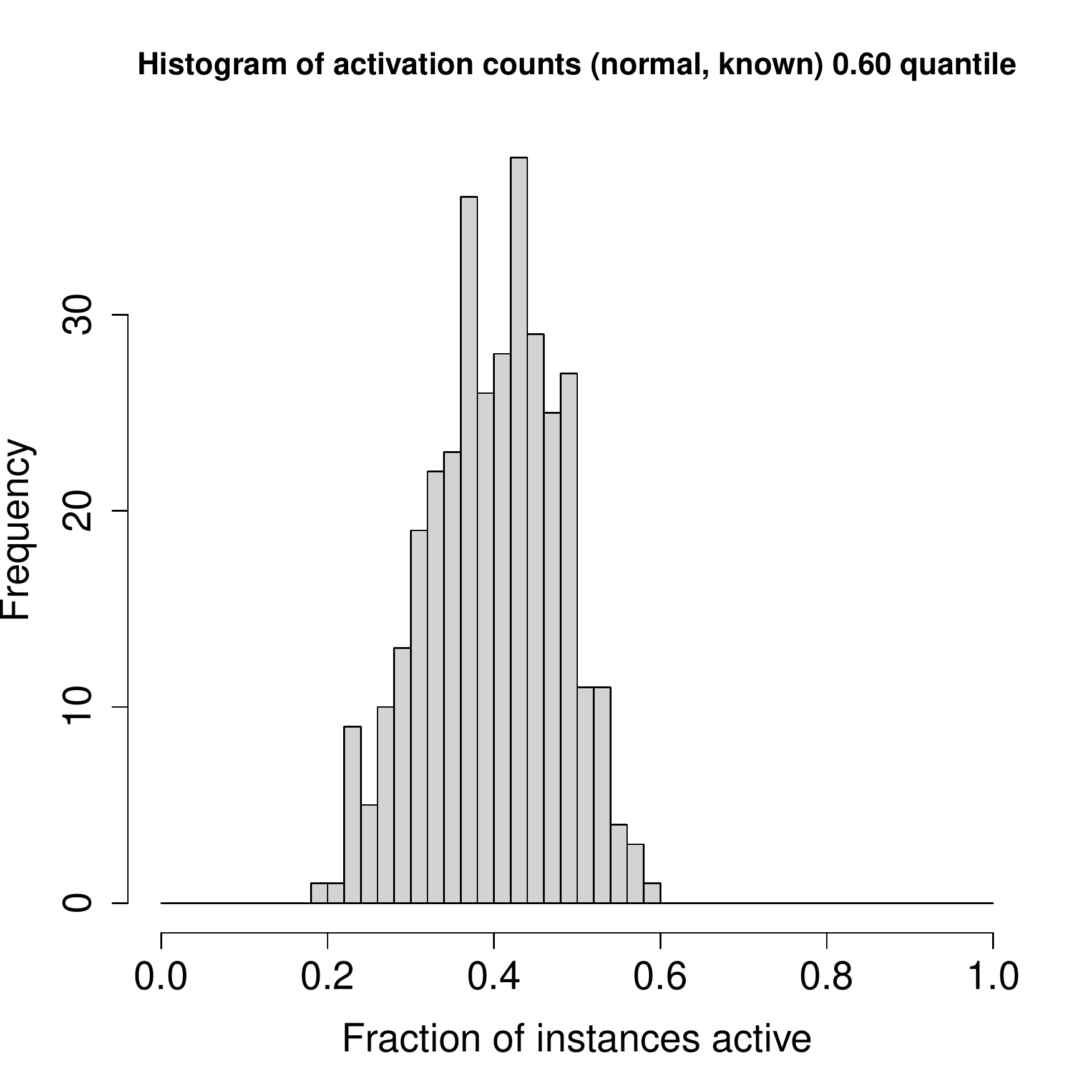}
    \includegraphics[width=3in]{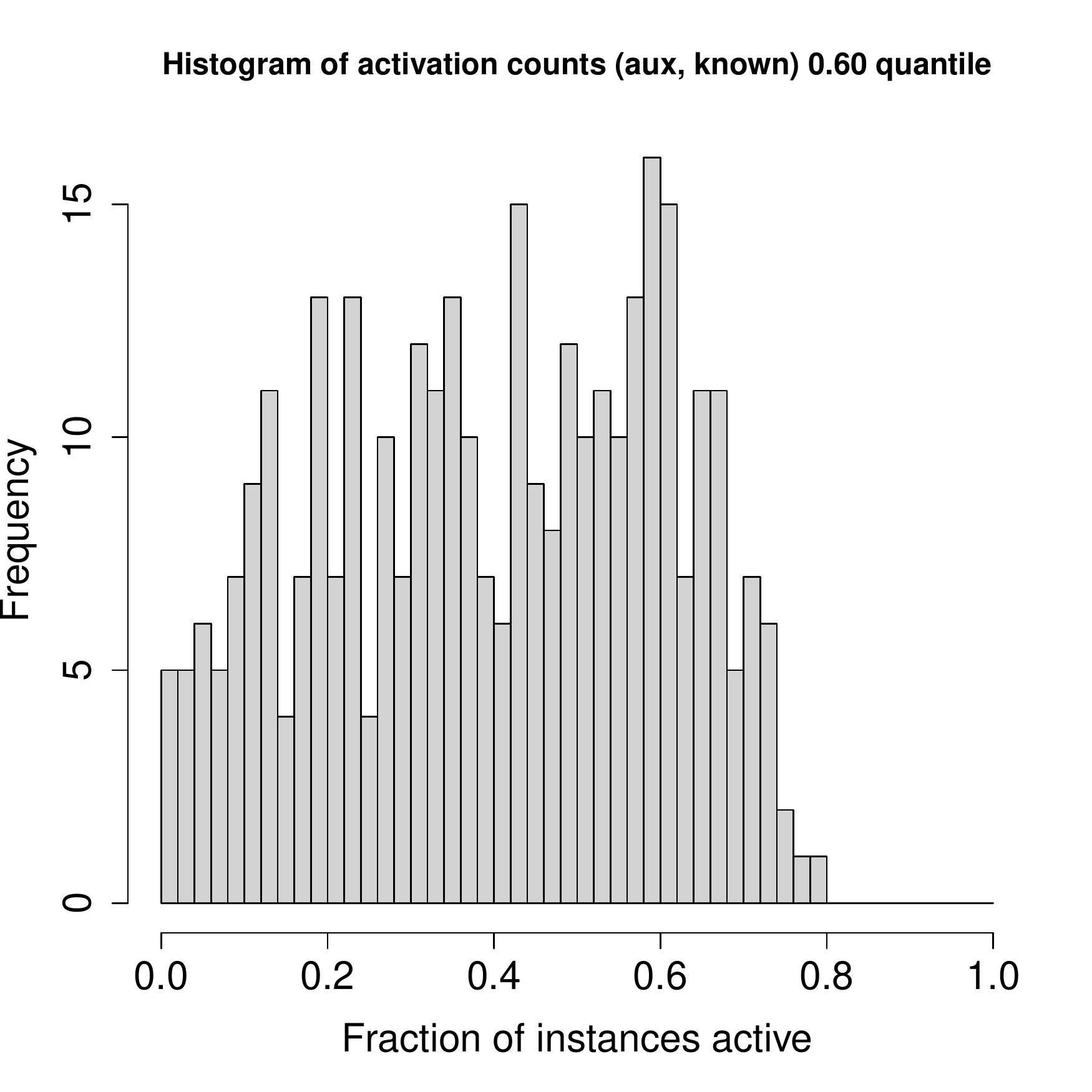}
    \caption{Fraction of instances in which $z_{ij} \geq \theta$ for $\theta$ set to the 0.60 quantile of the activations. Left: Normal training; Right: Auxiliary training.}
    \label{fig:activation-count-histograms}
\end{figure}

It is interesting to ask how the activity of each feature is distributed across the 1000 known-class test set images. One might expect that with Auxiliary training, features would need to become more specialized to particular classes or subsets of classes compared to Normal training. To explore this, we set the threshold $\theta$ to correspond to the $q$th quantile of the known-class activations and then computed the fraction of known-class test images $i$ where the activation $z_{ij}$ for each feature $j$ exceeded $\theta$. We set $\theta$ separately for Normal versus Auxiliary training, because the average magnitude of activations increases from 0.3394 for Normal training to 0.5118 for Auxiliary training. We plot the resulting 384 values (one per feature) as a histogram. If $z_{ij} \geq \theta$, we will say that feature $j$ is active on instance $i$. Figure~\ref{fig:activation-count-histograms} plots Normal and Auxiliary activation histograms for $q=0.60$. We observe that under Normal training, most features are active on 30-50\% of the instances, and no features are active on less than 20\% or more than 65\% of the instances. In contrast, under Auxiliary training, the distribution becomes more uniform. There are many more features that are active less often. At the extreme low end, there are 5 features that are only active on 0-2\% of the instances. This confirms our expectation that many features have become more specialized. However, there are also a few features that are active more often, with 20 features active on at least 70\% of the instances. Further study is needed to understand how the classifier is using these highly-active features.

This analysis shows that Prediction 3 is also correct, and it suggests a general strategy for improving classifiers and anomaly detectors. Previous work has used auxiliary labeled data for pre-training. For example, it is standard practice to pretrain on Imagenet-1K and then fine-tune the resulting network on the $K$ classes of interest. Our approach to auxiliary training instead trains on the combination of the $K$ classes of interest and all of the auxiliary classes. Auxiliary training is also related to Outlier Exposure \citep{Hendrycks2019}, which employs unlabeled novel-class data to maximize the entropy of $P(y|x)$. 

\section{Discussion}

In retrospect, it is not surprising that the standard model for open set detection works by detecting the lack of familiarity rather than the presence of novelty. There is some evidence that mammalian visual systems exhibit a similar limitation. In a famous experiment,  \cite{Blakemore1970} raised kittens (starting at age 2 weeks and continuing until they were 5 months old) in visual environments that consisted either of only black-and-white vertical lines or only black-and-white horizontal lines. This resulted in permanent changes to their visual perception. 
\begin{quote}
    There were [\ldots] differences between cats reared in horizontal and vertical environments. They were virtually blind for contours perpendicular to the orientation they had experienced. [\ldots] The differences were most marked when two kittens, one horizontally and the other vertically experienced, were tested simultaneously with a long black or white rod. If this was held vertically and shaken, the one cat would follow it, run to it, and play with it. Now if it was held horizontally, the other cat was attracted and its fellow completely ignored it.
\end{quote}
Perhaps our deep networks are similarly limited to only detecting variation that was present during training.

Auxiliary training provides richer variation during training, so we might expect that this would discover features that can represent the novel-class images better. However, Figures~\ref{fig:cifar-100-activations} and \ref{fig:cifar-100-class-5-contributions} show that the novel-class images still have significantly lower activations. It is therefore no surprise that methods that measure distances between $z_i$ vectors in the latent space still do not perform as well as the Max Logit method. For example, on CIFAR-100 with Auxiliary training, the Mahalanobis method only achieves an AUROC of 0.588$\pm$0.038, compared to 0.886$\pm$0.036 for Max Logit. The key benefit of the Max Logit approach is that it considers, separately for each class $k$, only the features relevant to that class. This turns out to be a potent anomaly detection method, and we speculate that the Max Logit strategy will continue to work well even in cases where the learned features can represent the novel-class instances well. 

How does the standard model based on the Max Logit method compare to techniques such as ODIN \citep{liang2018enhancing}, Generalized ODIN \citep{hsu2020generalized}, and the Adversarial Reciprocal Points method \citep{Chen2021}? ODIN is a variation on Max Softmax that introduces a perturbation to the image to try to increase the max softmax value. It then computes a temperature-rescaled Max Softmax novelty score. The underlying hypothesis is that it is easier (for a gradient ascent optimizer) to improve the softmax scores of known-class images than it is to increase the scores of novel-class images. However, we have not observed this in our experiments, and the ODIN authors did not present any direct evidence to support this claim. In any case, ODIN, because it is only a minor change to Max Softmax, still relies on measuring familiarity rather than novelty.

Generalized ODIN replaces the logit computation of the standard model with a ratio of two learned functions: $h_k(z)/g(z)$. The $h_k(z)$ functions (which are learned separately for each class $k$) can be viewed as more flexible versions of the logit scores, and $g(z)$ can be viewed as providing a learned temperature value. In our experience, we found that $h_k$ gave the best novelty detection scores, and these were better than the Max Softmax scores. Replacing the logit score $w_h^{\top} z$ with $h_k(z)$ does not change the fact that the learned network is measuring familiarity rather than novelty. 

One of the strongest alternatives to the standard model is the Adversarial Reciprocal Points (ARPL+CS) method of \cite{Chen2021}. It seeks to induce greater separation in the latent space between the known classes by introducing two innovations: (a) improving the representations for the data points in each class by maximizing their latent-space distance from a class-specific ``reciprocal point'' and (b) using a GAN to generate pseudo-novelties in the ``center'' of the latent space and then training discriminatively against them. Both of these mechanisms can be viewed as enhancing familiarity-based novelty scoring by pulling the known-class $z$ vectors further away from the center of the space where the novel-class $z$ vectors will be located.

Much prior work on open set and out-of-distribution detection has framed the problem as one of uncertainty quantification. The key hope is that the learned classifiers will be more uncertain when shown novel-class or out-of-distribution images. While the Max Logit strategy can be interpreted in this way, we believe that our focus on representation leads to more fruitful directions for future work, including the auxiliary training idea presented here. The uncertainty perspective motivates most work in Bayesian and ensemble approaches to deep anomaly detection. However, our view is that modeling epistemic uncertainty over a familiarity-based feature representation will yield only modest improvements. The FH suggests that ensemble methods need to explore the space of \textit{representations} in order to gain substantial anomaly detection improvements.

In Prediction 2, we raised one weakness of familiarity as a novelty detection mechanism---namely that occlusion that removes familiarity will lead to false anomaly detections. Adversarial attacks provide another threat to familiarity-based novelty detection. By applying existing attack algorithms (e.g., the FSGM \cite{Goodfellow2014}), we predict that it will be very easy to raise the logit score of at least one class and thereby hide a novel class image from novelty detection. It may also be possible to depress the logit scores of enough classes to create false anomaly alarms.

Another scenario in which familiarity-based novelty detection will fail is when the image contains both a familiar object and a novel object. Because the familiar object will activate all of the familiar features, the max logit score of the image will be high, and the novel object will go undetected. If novel and known objects are likely to co-occur in an image, a reasonable strategy would be to first perform object detection and then score each detected object separately. 

\section{Concluding Remarks}
This paper has presented evidence to support the hypothesis that existing state-of-the-art deep anomaly detection methods for computer vision operate by detecting the lack of familiarity rather than the presence of novelty. This appears to be an inherent consequence of learning representations from training data that does not contain the novel classes. 

One possible path for overcoming the limitations of familiarity-based anomaly detection would be to learn detectors of generic ``interesting image content''. If such a detector finds a region of interest in the image that does not activate the expected number of learned features, that could be a signal that the region contains novel objects. Exploring such research directions should be a high priority for deep anomaly detection research.

\section*{Acknowledgements}
The information provided in this paper is derived from two efforts sponsored by the Defense Advanced Research Projects Agency (DARPA) and awarded to Raytheon BBN Technologies under Contract Number HR001120C0022 and to SRI International under Contract Number HR001119C0112. Any opinions, findings and conclusions or recommendations expressed in this paper are those of the authors and do not necessarily reflect the views of the DARPA.

The authors thank our colleagues Fuxin Li, Stefan Lee, Alan Fern and the Oregon State Robust AI research group for helpful discussions and advice. We also thank the anonymous reviewers for asking questions that led us to improve the paper.

\clearpage
%
\bibliographystyle{plainnat}
\bibliography{familiarity}
\end{document}